\documentclass[10pt,twocolumn]{article}

\usepackage[letterpaper,margin=0.72in,columnsep=0.24in]{geometry}
\usepackage[T1]{fontenc}
\usepackage[utf8]{inputenc}
\usepackage{times}
\usepackage{microtype}
\usepackage{graphicx}
\usepackage{booktabs}
\usepackage{amsmath,amssymb}
\usepackage{array}
\usepackage{tabularx}
\usepackage{multirow}
\usepackage{url}
\usepackage[hidelinks]{hyperref}
\usepackage[superscript]{cite}
\usepackage{caption}
\usepackage{dblfloatfix}
\usepackage{float}
\usepackage{titlesec}
\usepackage{textcomp}

\titleformat{\section}{\large\bfseries}{\thesection}{0.5em}{}
\titleformat{\subsection}{\normalsize\bfseries\itshape}{\thesubsection}{0.5em}{}
\titlespacing*{\section}{0pt}{8pt}{3pt}
\titlespacing*{\subsection}{0pt}{6pt}{2pt}

\title{\bfseries Forecasting the Winner of a Live Tennis Match}
\author{%
\begin{minipage}[t]{0.44\textwidth}
\centering
Charles Xie\\[-1pt]
\small\textit{Student}\\[2pt]
\small Natick High School\\
\small 15 West Street, Natick, Massachusetts 01760, USA\\
\small \href{mailto:charlesxie157@gmail.com}{charlesxie157@gmail.com}
\end{minipage}
\hfill
\begin{minipage}[t]{0.44\textwidth}
\centering
Aneesh Muppidi\\[-1pt]
\small\textit{Mentor}\\[2pt]
\small University of Oxford\\
\small Oxford, United Kingdom\\
\small \href{mailto:aneesh.muppidi@magd.ox.ac.uk}{aneesh.muppidi@magd.ox.ac.uk}
\end{minipage}}
\date{}

\begin{document}
\maketitle

\begin{abstract}
With the rise of live sports betting in recent years, tennis forecasting has expanded from pre-match prediction to models that update win probabilities as a match unfolds. A central challenge in creating such a model is the constant need for models to adapt to score and performance changes. This study examines how pre-match and live information can be most effectively integrated into a model to produce accurate win-probability estimates. The analysis uses 8,222 Grand Slam matches containing a total of 1,505,355 points. Five models were evaluated using a chronological split, with matches from 2011--2021 used for training, 2022 for validation, and 2023--2024 for testing. Trace, a hybrid model, achieved accuracies of 76.06\%, 82.15\%, and 88.34\% at 25\%, 50\%, and 75\% match progress, suggesting that hybrid modeling is a practical approach to live tennis forecasting.
\end{abstract}

\section{Introduction}

In recent decades, advances in real-time data collection and digital betting platforms have contributed to the rapid growth of live sports betting.\cite{ding2025} Consequently, live betting has developed into a substantial market, with tennis emerging as a major contribution to its growth. In fact, approximately 80\% of the money wagered on tennis matches has been reported to be placed in-play.\cite{sipko2015}

Unlike pre-match forecasting, live tennis prediction provides an influx of live statistics. This can be seen with the growing use of technologies such as Hawk-Eye\textsuperscript{\textregistered}, substantially augmenting the amount of shot-level and contextual information that can be collected throughout a tennis match.\cite{fitzpatrick2024} Although this information may contain useful predictive signals, it can also be harmful if incorporated indiscriminately. For instance, too little or too much live data can trick a model into thinking in a false direction.\cite{gollub2021} Thus, a successful live model must determine a way to balance the incorporation of live data.\cite{ciobanu2023}

\subsection{Related Work}

Klaassen and Magnus introduced a hierarchical Markov model that updates a player's win probability from the current score.\cite{klaassen2003} Gollub extended hierarchical Markov prediction by adding serve probabilities from pre-match data and updating them using a beta-binomial model mid-match. For live matches, this model reached a maximum accuracy of 76.5\% across all points.\cite{gollub2017}

Lerner, Badri, and Monogue built on Gollub's model by introducing DeepTennis, a long short-term memory (LSTM) model with the same goal as Gollub's work. Their model combined sequential point-by-point match data with pre-match probability estimates, achieving an overall test accuracy of 79.5\%.\cite{lerner2019} Among the live tennis studies reviewed, this was the highest reported accuracy. However, because this model used different test years and data, it serves as a contextual benchmark rather than a directly comparable baseline for this study.

Although this study focuses primarily on live-match prediction, pre-match models remain valuable because they provide predictive information before the match starts, allowing a model to start away from a 50/50 prediction. One prominent and useful model is the creation of an Elo system for tennis, which has been explored in depth by multiple papers. Vaughan Williams et al. found that Elo-based methods generally provided stronger forecasts of current player performance than official rankings, supporting their use as pre-match probability estimates.\cite{vaughan2021} More concretely, Kovalchik compared eleven pre-match tennis forecasting approaches and found that the career-history Elo model performed second best, just below the market consensus odds, achieving 70\% accuracy overall.\cite{kovalchik2016} This further corroborates the use of Elo as a reliable pre-match prior.

This study investigates different methods of adapting pre-match and live tennis data to produce a strong live win-probability estimate. Unlike the prior studies discussed above, it compares score structure, Elo, live serve updates, and machine learning within the same chronological evaluation.

\section{Data}

This study uses Jeff Sackmann's Grand Slam point-by-point data along with ATP and WTA match results.\cite{sackmann_atp,sackmann_slam,sackmann_wta} The Grand Slam data provide live information recorded throughout each match, including the server, point winner, current score, serve outcomes, rally length, and break-point information. The ATP and WTA match data are used separately to calculate each player's pre-match Elo rating.

\subsection{Data Preprocessing}

After data cleaning, the dataset retained 8,222 matches and 1,505,355 point-level prediction states. Table~\ref{tab:data-audit} shows the division of data and other key data characteristics. These counts refer to the dataset before filtering by Elo availability.

\begin{table}[H]
\centering
\caption{Composition, filtering, and Elo coverage of the prepared dataset; Association of Tennis Professionals (ATP) and Women's Tennis Association (WTA) data are included, and Elo ratings were matched to 96.0\% of prepared matches. Source: authors' analysis of the cited datasets.\cite{sackmann_atp,sackmann_slam,sackmann_wta}}
\label{tab:data-audit}
\small
\begin{tabular}{lrr}
\toprule
\textbf{Quantity} & \textbf{Matches} & \textbf{Point states} \\
\midrule
\multicolumn{3}{l}{\emph{Prepared sample}} \\
Total & 8,222 & 1,505,355 \\
ATP & 4,181 & -- \\
WTA & 4,041 & -- \\
\addlinespace
\multicolumn{3}{l}{\emph{Temporal split}} \\
Training, 2011--2021 & 6,774 & 1,238,084 \\
Validation, 2022 & 473 & 87,439 \\
Testing, 2023--2024 & 975 & 179,832 \\
\addlinespace
\multicolumn{3}{l}{\emph{Filtering and coverage}} \\
Raw matches excluded (21.8\%) & 2,286 & -- \\
Matched to Elo ratings (96.0\%) & 7,896 & -- \\
\bottomrule
\end{tabular}
\end{table}

A match was kept only if the following criteria were met: it produced a valid, non-tied final set score, the winner had completed either two or three sets, more than 20 points were recorded, and the server and point winner were identified as either player 1 or player 2. This process removed nonessential data, such as matches with walkovers.

After filtering, the data were organized so that each row shows the state of the match before a point was played. Each row included the current set, game, and point scores, the server, the match format, and other live features. The eventual match winner was stored in a separate column as the target variable, ensuring that each prediction is based only on information available at that point in the match. This was an essential step to mitigate data leakage.

Next, the match states were converted into a numerical format that the models could process. For instance, tennis point scores had to be encoded numerically because the Markov recursion cannot directly use labels such as 15, 30, 40, and advantage. Table~\ref{tab:score-encoding} shows the encoding used.

\begin{table}[H]
\centering
\caption{Numerical encoding of tennis point scores, with advantage encoded as 4. Source: authors' implementation.\cite{xie2026}}
\label{tab:score-encoding}
\begin{tabular}{cc}
\toprule
\textbf{Raw point score} & \textbf{Encoded value} \\
\midrule
0 & 0 \\
15 & 1 \\
30 & 2 \\
40 & 3 \\
AD & 4 \\
\bottomrule
\end{tabular}
\end{table}

The other match-state variables, such as server identity and tiebreak status, were similarly standardized.

After encoding the match states, new cumulative features were constructed from the raw point-by-point data from the dataset. Unlike raw features, such as an ace or point won, these newly constructed features, such as serve rates and game or set outcomes, better capture a player's overall performance. Table~\ref{tab:constructed-features} summarizes the main groups of features used in the processed dataset.

\begin{table}[H]
\centering
\caption{Principal variable groups in the processed point-level dataset, summarizing score state, live serve performance, rally and ace context, and derived match state. Source: authors' processing of the Grand Slam point-by-point data.\cite{sackmann_slam}}
\label{tab:constructed-features}
\scriptsize
\setlength{\tabcolsep}{2pt}
\begin{tabularx}{\columnwidth}{>{\raggedright\arraybackslash}p{0.18\columnwidth} >{\raggedright\arraybackslash}p{0.32\columnwidth} >{\raggedright\arraybackslash}X}
\toprule
\textbf{Feature group} & \textbf{Examples} & \textbf{Description} \\
\midrule
Score state & \texttt{best\_of, p1\_sets, p2\_sets, p1\_games, p2\_games, p1\_score, p2\_score, p1\_serving, tiebreak} & Describes the match format and the current set, game, point-score, and server state. \\
Match outcome & \texttt{winner, y} & The final match result is stored and used exclusively as the prediction target. \\
Live serve performance & \texttt{p1\_serve\_rate, p2\_serve\_rate, p1\_serve\_n, p2\_serve\_n} & Tracks each player's observed serve-point win rate and corresponding sample size, before the current point. \\
Rally and ace context & \texttt{rally\_avg, recent\_rally\_avg, p1\_ace\_rate, p2\_ace\_rate} & Summarizes rally length and ace performance from previously completed points when such measurements are available. \\
Derived state summaries & \texttt{sets\_diff, games\_diff, score\_diff, pts\_played} & Provides compact representations of the evolving match state from player 1's perspective. \\
\bottomrule
\end{tabularx}
\end{table}

\subsection{Elo}

To represent player strength before each match, Elo ratings were built from historical ATP and WTA results. Mirroring the Elo approach of Kovalchik,\cite{kovalchik2016} every player started with an Elo of 1500. Ratings were then updated in chronological order, with the size of each update decreasing as a player played more matches.

For a match between players $i$ and $j$, player $i$'s expected probability of winning was
\begin{equation}
q_{i,t}=\frac{1}{1+10^{(R_{j,t}-R_{i,t})/400}},
\end{equation}
where $R_{i,t}$ and $R_{j,t}$ are the players' ratings immediately before the match. After the outcome was observed, the rating was updated as
\begin{equation}
R_{i,t+1}=R_{i,t}+K_{i,t}(S_{i,t}-q_{i,t}),
\qquad
K_{i,t}=\frac{250}{(m_{i,t}+5)^{0.4}},
\end{equation}
where $S_{i,t}\in\{0,1\}$ gives the result of the match and $m_{i,t}$ is the number of matches player $i$ has completed beforehand.

Each player's Elo rating was recorded before the outcome of the current match, preventing information from the match itself from interfering with the pre-match Elo estimate. These ratings were then added to the Grand Slam point-by-point data.

\subsection{Evaluation Split}

The data were split chronologically to prevent data leakage and mirror how a model would work in a real-time environment.

Matches from 2011 to 2021 were used for training, 2022 was used for validation and parameter tuning, and matches from 2023--2024 were used for testing.

\section{Methods}

To examine the value that different sources of information add to live tennis predictions, this paper evaluates five models: a symmetric Markov baseline, an Elo-asymmetric Markov model, a serve-shrink Markov model, a histogram gradient-boosting model (HGBM), and Trace. The models build on one another in different ways, making it possible to compare the value of the different methodologies.

\subsection{Markov Score Recursion}

The Markov recursion forms the backbone of every model except the HGBM. Its role is to take point-level serve probabilities and use tennis scoring rules to calculate the probability that player 1 eventually wins the match.

Markov recursion is a natural fit for tennis because of the way its scoring system is organized: points make up games, games make up sets, and sets determine the match.\cite{omalley2008} From any score, the model can consider the two possible outcomes of the next point and calculate how each outcome changes the probability of eventually winning the match.

Let $p_1$ be player 1's probability of winning a point on serve and $p_2$ be player 2's probability of winning a point on serve. The probability that player 1 wins the next point is
\begin{equation}
q=
\begin{cases}
p_1, & \text{when player 1 serves},\\
1-p_2, & \text{when player 2 serves}.
\end{cases}
\end{equation}

More formally, the complete match state can be represented as
\begin{equation}
s=(S_1,S_2,G_1,G_2,P_1,P_2,\sigma,\tau,b),
\end{equation}
where $S_1$ and $S_2$ are the sets won by each player, $G_1$ and $G_2$ are the games won in the current set, $P_1$ and $P_2$ are the encoded point scores, $\sigma$ identifies the server, $\tau$ indicates whether the current game is a tiebreak, and $b$ is the best-of match format. Let $\mathcal{T}_1(s)$ and $\mathcal{T}_2(s)$ denote the states reached when player 1 or player 2 wins the next point. The probability that player 1 eventually wins the match is then
\begin{equation}
V(s)=q(s)V\!\left(\mathcal{T}_1(s)\right)+\left(1-q(s)\right)V\!\left(\mathcal{T}_2(s)\right).
\end{equation}

If $m(b)=\lceil b/2\rceil$ is the number of sets required to win the match, the terminal conditions are
\begin{equation}
V(s)=
\begin{cases}
1, & S_1=m(b),\\
0, & S_2=m(b).
\end{cases}
\end{equation}

Deuce and tiebreak states require slight modifications to the standard recursion. From deuce, player 1's probability of winning the game can be written as
\begin{equation}
D(q)=\frac{q^2}{q^2+(1-q)^2},
\end{equation}
where $q$ is player 1's probability of winning the next point. Tiebreaks use the same recursive framework, but the server changes according to the standard service sequence,
\begin{equation}
1,2,2,1,1,2,2,\ldots,
\end{equation}
with one player serving the opening point followed by alternating two-point service turns.

\subsection{Symmetric Markov Baseline}

The symmetric Markov baseline treats both players as equally strong:
\begin{equation}
p_a=p_b=p.
\end{equation}

The value of $p$ is selected using the validation season from $\{0.60,0.61,0.62,0.63,0.64,0.65\}$. At 25\%, 50\%, and 75\% match progress, the selected values were 0.65, 0.64, and 0.60, respectively. Since both players receive the same serve probability, this model receives information solely from the match score. For instance, if two matches have the same score state, this model gives them the same win probability regardless of who the players are.

This baseline was created to isolate the predictive value of tennis scoring.

\subsection{Elo-asymmetric Markov Model}

Because players often differ in strength, the underlying assumption behind the symmetric Markov model is not practical for real tennis matches. The Elo-asymmetric model resolves this issue by allowing the two players to start with different serve probabilities. The model uses the pre-match Elo difference:
\begin{equation}
d=\mathrm{Elo}_1-\mathrm{Elo}_2.
\end{equation}

Subsequently, the Elo gap is converted into a serve-probability edge:
\begin{equation}
e=\mathrm{clip}(\alpha d,-0.15,0.15),
\end{equation}
where $\alpha$ is a validation-tuned slope. The baseline serve probability was searched from 0.59 to 0.65 in increments of 0.01. The slope grid was $4,6,9,13,18,$ and $22\times10^{-5}$. At 25\%, 50\%, and 75\% match progress, the selected $(\beta,\alpha)$ values were $(0.64,1.3\times10^{-4})$, $(0.59,1.3\times10^{-4})$, and $(0.59,1.3\times10^{-4})$, respectively.

The serve probabilities are then
\begin{equation}
p_a=\mathrm{clip}(\beta+e,0.45,0.88),
\end{equation}
\begin{equation}
p_b=\mathrm{clip}(\beta-e,0.45,0.88),
\end{equation}
where $\beta$ is the tour-average baseline serve probability, and the clipping bounds were added to prevent the model from producing unrealistic serve probabilities.

Overall, this model can prevent the initial match estimate from defaulting to 50--50 when the players differ in strength.

\subsection{Serve-shrink Markov Model}

The Elo-asymmetric model assigns serve probabilities using pre-match information but does not update them as the match progresses. Suppose a player with a historical serve-point win rate of 70\% wins 9 of their first 10 serve points, thus producing a live serve rate of 90\%. Although this may indicate that the player is playing especially well, the estimate is based on a limited sample and may reflect a temporary variation in skill level, and it is unlikely that they will consistently play at this skill level throughout the entire match.

The serve-shrink model addresses this limitation by weighing the player's observed in-match serve rate against their pre-match serve rate. In this example, the player's estimated serve probability should increase, but not immediately to 90\%. The live serve rate is therefore weighted in proportion to the amount of in-match evidence available.

Using a simplified form of the Bayesian shrinkage approach described by Colin and Bechler,\cite{colin2015} the general equation for the serve-shrink model is
\begin{equation}
\hat{\theta}
=\frac{n}{n+\kappa}\hat{\theta}_{\mathrm{live}}
+\frac{\kappa}{n+\kappa}\theta_{\mathrm{prior}},
\end{equation}
where $\hat{\theta}_{\mathrm{live}}$ is the live estimate, $\theta_{\mathrm{prior}}$ is the pre-match estimate, $\hat{\theta}$ is the resulting updated estimate, $n$ is the number of observed serve points, and $\kappa$ is a prior pseudo-count that controls the amount of shrinkage.

Let $\pi_1$ and $\pi_2$ denote the Elo-based prior serve-point win probabilities for players 1 and 2. Let $r_1$ and $r_2$ denote the observed in-match serve-point win rates, and $n_1$ and $n_2$ denote the corresponding numbers of serve points observed before the current match state. The updated serve probabilities are
\begin{equation}
p_a=\frac{n_1r_1+\kappa\pi_1}{n_1+\kappa},
\qquad
p_b=\frac{n_2r_2+\kappa\pi_2}{n_2+\kappa}.
\end{equation}

The live serve rate receives weight $n_i/(n_i+\kappa)$ while the pre-match prior receives weight $\kappa/(n_i+\kappa)$. As $n_i$ increases, the weight progressively shifts toward the observed live serve rate.

The parameter $\kappa$ controls the strength of shrinkage toward the pre-match estimate, with large values giving more weight to the prior and smaller values giving more weight to recent serve performance. To select the value of $\kappa$, the values $\{40,80,160,320,640\}$ were compared using validation log loss. The final values were then determined to be 640, 160, and 40 at 25\%, 50\%, and 75\% match progress, respectively.

The resulting $p_a$ and $p_b$ values are then passed into the Markov recursion with the current score state to obtain player 1's match-win probability.

\subsection{Histogram Gradient-Boosting Baseline}

The histogram gradient-boosting model was used as a machine-learning baseline and was implemented with scikit-learn's HistGradientBoostingClassifier.\cite{pedregosa2011} It was designed to test the effectiveness of machine learning in detecting the winner of a match. Unlike the Markov-based models discussed earlier, it learns player 1's match-win probability solely from the current score and live features.

Let $z$ represent the score and live performance features available at the current match state. The baseline prediction is
\begin{equation}
\hat{p}_H=g_{\phi}(z),
\end{equation}
where $g_{\phi}$ is a histogram gradient-boosting classifier and $\hat{p}_H$ is the predicted probability that player 1 wins the match.

The HGBM provides a machine-learning baseline that estimates live win probability directly from live match features without using a Markov prediction.

Four hyperparameter settings were evaluated by validation log loss: $(0.03,150,15,1.0)$, $(0.03,300,15,1.0)$, $(0.05,200,15,3.0)$, and $(0.05,200,31,3.0)$. The entries correspond to learning rate, maximum iterations, maximum leaf nodes, and L2 regularization. Training used log loss with early stopping disabled and a fixed random state of 42. After tuning, the selected configuration was refit on the combined training and validation data and evaluated on the test set.

XGBoost and LightGBM were also evaluated using the same training and validation procedure.\cite{chen2016,ke2017} However, because the histogram gradient-boosting classifier achieved the lowest validation log loss, it was selected as the machine-learning baseline.

\subsection{Trace}

The final model used in this study is Trace. It combines predictions from the Elo-asymmetric and serve-shrink Markov models with the current score and additional live features. These inputs are passed to a histogram gradient-boosting model (HGBM), which produces the final match-win probability.

Let $\hat{p}_M(s)$ denote the Elo-asymmetric Markov probability, $\hat{p}_S(s)$ denote the serve-shrink Markov probability, and $z(s)$ denote the remaining score-state and live-match features. The resulting feature vector is
\begin{equation}
\mathbf{x}(s)=
\begin{bmatrix}
\hat{p}_M(s)\\
\hat{p}_S(s)\\
\mathrm{logit}(\hat{p}_M(s))\\
\mathrm{logit}(\hat{p}_S(s))\\
\hat{p}_S(s)-\hat{p}_M(s)\\
z(s)
\end{bmatrix},
\end{equation}
where
\begin{equation}
\mathrm{logit}(p)=\log\!\left(\frac{p}{1-p}\right).
\end{equation}

Trace then produces
\begin{equation}
\hat{p}_{\mathrm{Trace}}(s)=f_{\theta}(\mathbf{x}(s)),
\end{equation}
where $f_{\theta}$ is the HGBM.

The hypothesized advantage that Trace has over the Elo-asymmetric Markov and serve-shrink models is its ability to capture non-linearity. With the help of the HGBM, Trace was predicted to be able to more effectively capture nuanced and complex scenarios that the models could not. The vector $z(s)$ contains the remaining live-match features used by the HGBM. To prevent the possibility of data leakage, at every state $s$, these variables are constructed only from information available before the next point is played, preventing future information from entering the prediction.

\begin{figure}[H]
\centering
\includegraphics[width=\columnwidth]{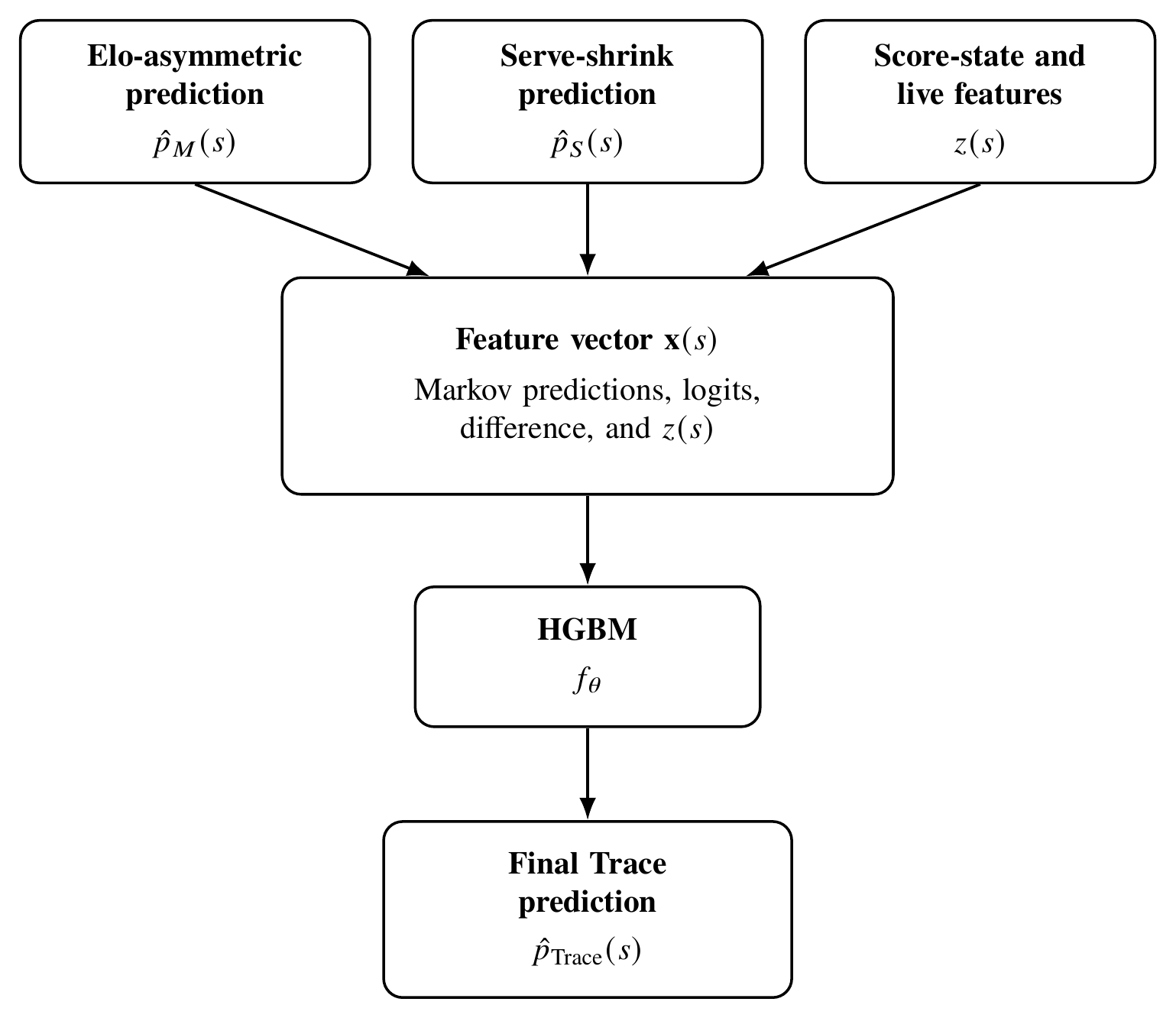}

\caption{Trace model pipeline.}
\label{fig:trace-pipeline}
\end{figure}

The Trace pipeline is shown in Figure~\ref{fig:trace-pipeline}.

Hyperparameters were chosen according to validation log loss. Four HGBM settings were evaluated: $(0.03,7,1.0)$, $(0.03,15,1.0)$, $(0.05,7,1.0)$, and $(0.05,15,3.0)$, with each tuple giving the learning rate, maximum number of leaf nodes, and L2 regularization. All configurations used log loss, early stopping, and a maximum of 250 boosting iterations. The 25\%, 50\%, and 75\% checkpoint models used random states 19, 30, and 6, respectively.

\section{Results and Discussion}

The models were evaluated on the test set, which contains matches strictly from 2023 and 2024. The accuracies of the models were collected at fixed intervals of match progress. For example, at 25\% match progress, the model received the score state and live features from approximately one-quarter of the way through each match. This was repeated at 50\% and 75\%. This setup measures how performance changes as more information about the match is revealed to the models. Additionally, the accuracies of the models over all points were examined.

The project reports two metrics: log loss and accuracy. These metrics are explained in detail in the following sections.

\subsection{Accuracy}

Accuracy is a widely used and intuitive measure of model performance, making it useful for straightforward comparisons between models. Table~\ref{tab:accuracy} reports the checkpoint and all-point results.

\begin{table}[H]
\centering
\caption{Model accuracy at fixed match-progress checkpoints and across all points; HGBM denotes the histogram gradient-boosting model, and Trace has the highest accuracy in every reported column. Source: authors' model evaluation.\cite{xie2026}}
\label{tab:accuracy}
\resizebox{\columnwidth}{!}{%
\begin{tabular}{lcccc}
\toprule
\textbf{Model} & \textbf{25\%} & \textbf{50\%} & \textbf{75\%} & \textbf{All Points} \\
\midrule
Symmetric Markov & 0.6851 & 0.7703 & 0.8544 & 73.21\% \\
Elo-asymmetric Markov & 0.7575 & 0.8050 & 0.8648 & 77.56\% \\
Serve-shrink Markov & 0.7564 & 0.7946 & 0.8720 & 77.68\% \\
HGBM & 0.6944 & 0.7918 & 0.8738 & 73.98\% \\
\textbf{Trace} & \textbf{0.7606} & \textbf{0.8215} & \textbf{0.8834} & \textbf{77.84\%} \\
\bottomrule
\end{tabular}
}
\end{table}

\begin{figure}[H]
\centering
\includegraphics[width=\columnwidth]{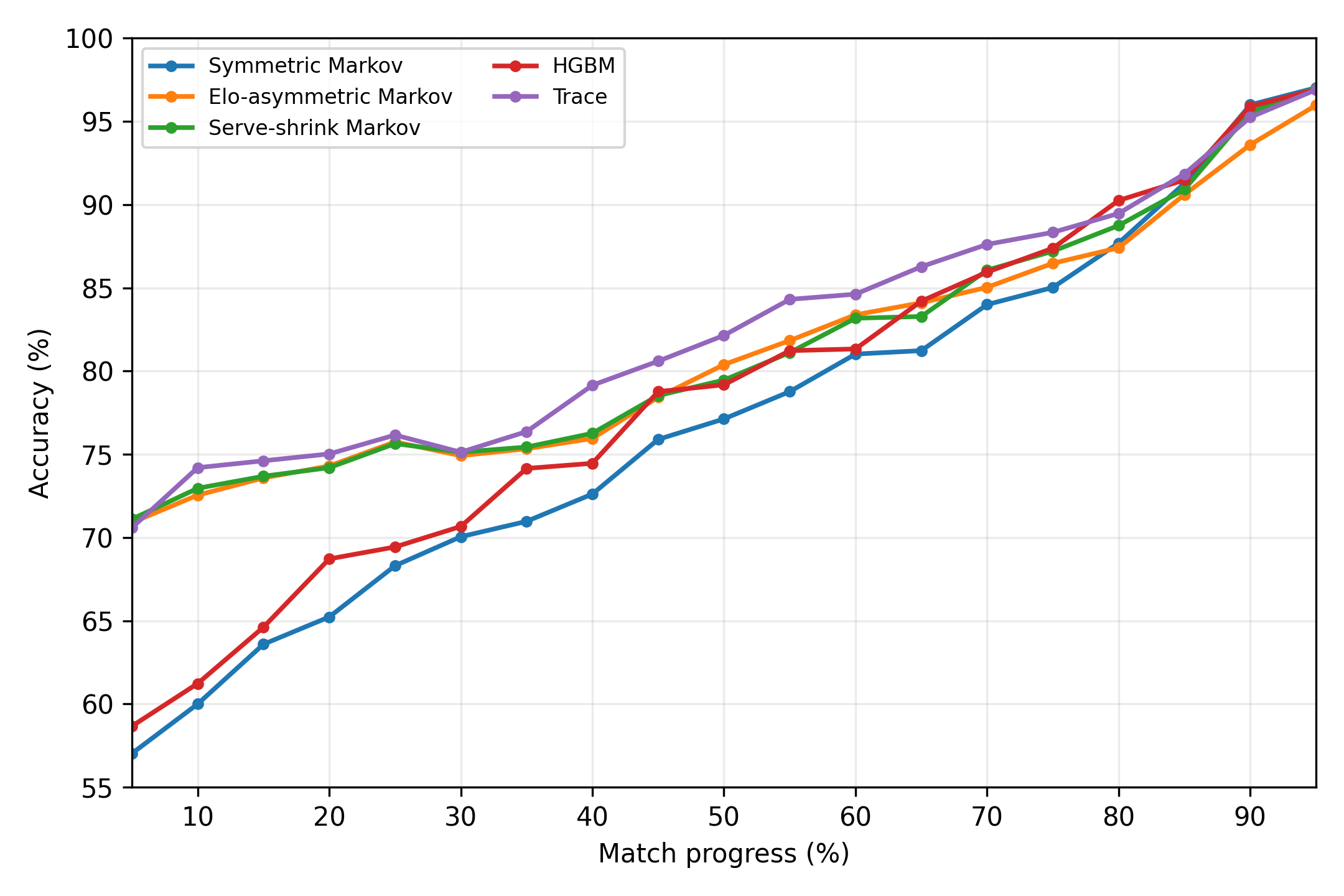}
\caption{Accuracy of the five models across match progress.}
\label{fig:model-accuracy}
\end{figure}

Figure~\ref{fig:model-accuracy} shows all five models across match progress.

Trace achieved the highest accuracy at each reported checkpoint and across all point-level predictions. Its largest advantage over the other models occurred during the middle stages of the match, particularly between approximately 40\% and 70\% match progress, as shown in Figure~\ref{fig:model-accuracy}.

Accuracy increased for all models as the match progressed, and the differences between the five plotted models narrowed near the end of the match. However, because accuracy does not assess the quality or confidence of the predicted probabilities, the models were also evaluated using log loss and calibration.

\subsection{Log Loss and Calibration}

To get a better grasp of model performance, the models' log losses were also calculated. Log loss evaluates the quality of the model estimates rather than only whether the predicted winner is correct. In essence, it rewards confident and correct estimates while penalizing confident incorrect predictions especially strongly. The loss function used for every model was binary log loss, which is defined as
\begin{equation}
L=-\frac{1}{N}\sum_{i=1}^{N}\left[y_i\log(p_i)+(1-y_i)\log(1-p_i)\right].
\end{equation}
Here, $p_i$ is the predicted probability that player 1 wins, $y_i\in\{0,1\}$ is the observed match outcome, and $N$ is the total number of predictions. A lower log loss indicates better probability estimates. Table~\ref{tab:logloss} reports log loss at the three checkpoints.

\begin{table}[H]
\centering
\caption{Model log loss at fixed match-progress checkpoints; HGBM denotes the histogram gradient-boosting model, and Trace has the lowest log loss at all three checkpoints. Source: authors' model evaluation.\cite{xie2026}}
\label{tab:logloss}
\resizebox{\columnwidth}{!}{%
\begin{tabular}{lccc}
\toprule
\textbf{Model} & \textbf{25\% Loss} & \textbf{50\% Loss} & \textbf{75\% Loss} \\
\midrule
Symmetric Markov & 0.6292 & 0.5363 & 0.3498 \\
Elo-asymmetric Markov & 0.5210 & 0.4549 & 0.3096 \\
Serve-shrink Markov & 0.5142 & 0.4266 & 0.2842 \\
HGBM & 0.5376 & 0.3788 & 0.2299 \\
\textbf{Trace} & \textbf{0.4753} & \textbf{0.3530} & \textbf{0.2002} \\
\bottomrule
\end{tabular}
}
\end{table}

Trace performed best across all evaluated match-progress checkpoints. To examine the reliability of its probability estimates more directly, the model was also evaluated using a calibration curve. Calibration curves complement log loss by comparing predicted win probabilities with observed win frequencies.

Following Guo et al.,\cite{guo2017} the predicted probabilities are divided into intervals, or bins, such as 0.60--0.75. For each bin $B_k$, the mean predicted probability is calculated as
\begin{equation}
\mathrm{Pred}_k=\frac{1}{|B_k|}\sum_{i\in B_k}p_i,
\end{equation}
and the corresponding observed win frequency is calculated as
\begin{equation}
\mathrm{Obs}_k=\frac{1}{|B_k|}\sum_{i\in B_k}y_i,
\end{equation}
where $p_i$ is the predicted probability that player 1 wins, $y_i$ is the observed match outcome, and $|B_k|$ is the number of predictions in bin $B_k$. A perfectly calibrated model has an observed win frequency equal to its mean predicted probability in each bin.

\begin{figure}[H]
\centering
\includegraphics[width=\columnwidth]{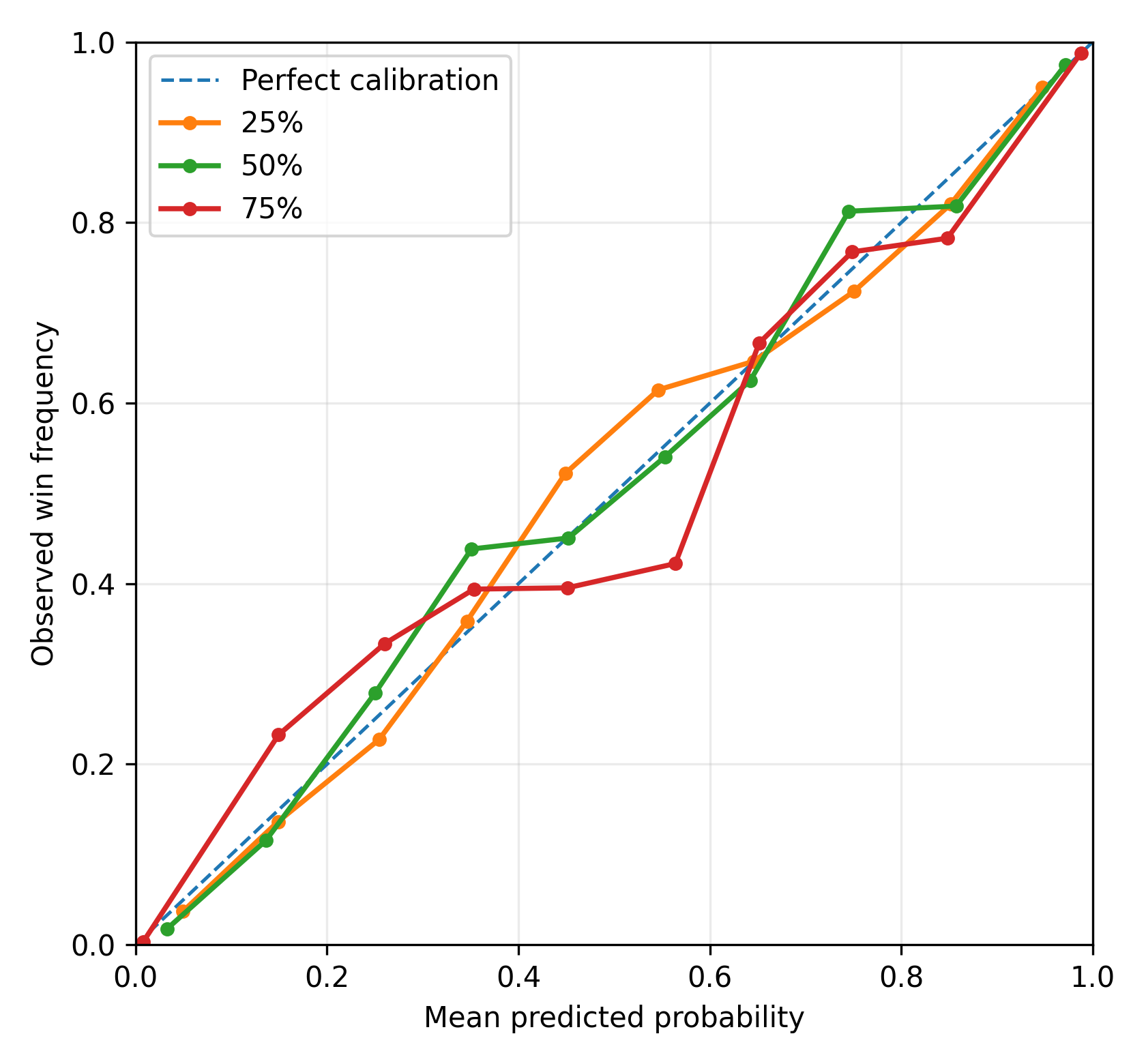}
\caption{Calibration curve for Trace at 25\%, 50\%, and 75\% match progress.}
\label{fig:calibration-curve}
\end{figure}

Figure~\ref{fig:calibration-curve} plots the mean predicted probability, $\mathrm{Pred}_k$, against the observed win frequency, $\mathrm{Obs}_k$, for each probability interval $B_k$. Predictions closer to the diagonal line $\mathrm{Obs}_k=\mathrm{Pred}_k$ indicate better calibration.

The reliability diagram indicates that Trace is reasonably well calibrated, especially at 25\% and 50\% match progress.

\subsection{Tour-specific Performance}

\begin{figure}[H]
\centering
\includegraphics[width=\columnwidth]{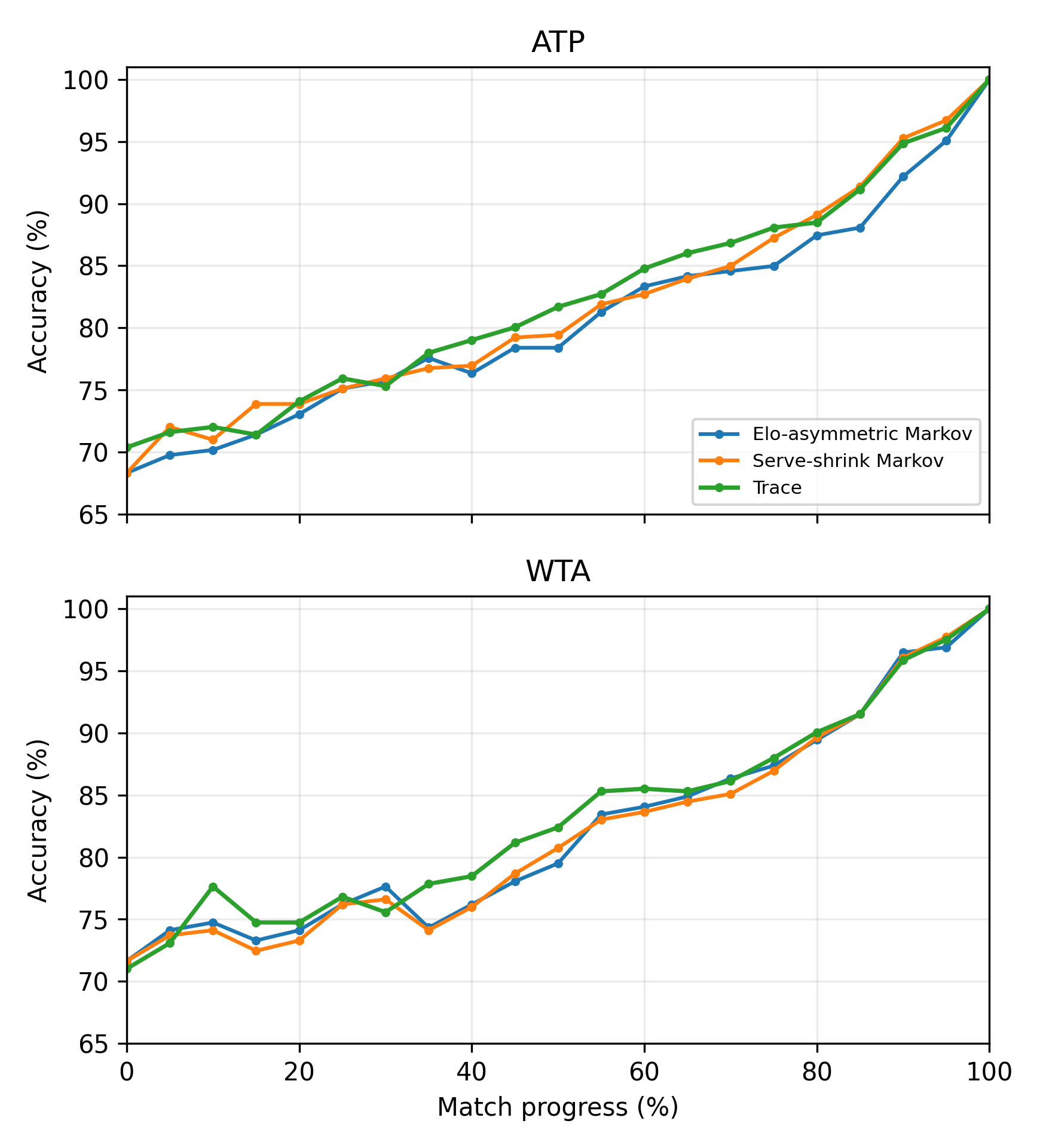}
\caption{Model accuracy by tour (ATP and WTA).}
\label{fig:tour-specific-accuracy}
\end{figure}

Figure~\ref{fig:tour-specific-accuracy} separates model accuracy by tour for ATP and WTA matches to observe if the two tours differed in model predictions. Such differences may be obscured when results from both tours are combined.

The main pattern is similar on both tours: Trace is almost always the strongest model across both ATP and WTA matches, especially through the middle portions of the match. It can also be observed that the WTA graph exhibits greater fluctuation than the ATP graph.

\subsection{Discussion}

The models developed along noticeably different performance trajectories as the match progressed, and those differences are useful to reveal what each approach is actually extracting from the available information. The most notable trajectory was the HGBM baseline: it was extremely weak early in matches, even falling behind the symmetric Markov baseline. Nonetheless, its performance improved substantially as matches unfolded. By 75\% match progress, it achieved the highest accuracy among the individual models and also produced lower log loss than all three Markov models at both 50\% and 75\%. This pattern suggests that as matches develop, live features become a substantially more valuable source of information for models than the Markov score recursion.

Another interesting performance was the serve-shrink model. On the surface, its accuracy gains over the Elo-asymmetric model were relatively small and were not consistent at every checkpoint. However, its distinction appeared in the log loss, where it consistently outperformed the Elo-asymmetric model. This finding suggests that the live serve update mainly improves the quality of the probability estimate rather than frequently changing which player the model expects to win. For instance, in some scenarios both models remained on the same side of the 0.5 threshold, but the serve-shrink model produced better-calibrated probability estimates; that is, it assigned higher confidence when correct and lower confidence when incorrect more often than the Elo-asymmetric model. While accuracy is still the sharpest way to evaluate model performance, this distinction still remains valuable because it allows users to trust the model's predictions better.

As expected, Trace produced the strongest overall results. As an ensemble of previous models, Trace was able to pool the advantages of all the models together and limit their weaknesses. For instance, the serve-shrink Markov outperformed the HGBM throughout the early stages of the match, but was outperformed by it during the later stages. Hence, because Trace utilizes both of these models, it was able to dominate throughout the entire match.

The tour-specific curves revealed similar model trajectories, but one noteworthy finding in this experiment was that the WTA predictions fluctuated more sharply over the course of a match than the ATP predictions. A plausible explanation is the smaller average serving advantage in women's tennis. Previous research has found that male Grand Slam players serve at higher velocities, produce more aces and unreturned serves, and win a greater proportion of service points than female players.\cite{reid2016} Professional men have also been reported to hold approximately 79\% of service games, with a substantially lower hold rate of 66\% for women.\cite{goff2025} Because breaking serves occur far more frequently in the WTA, the direction of a match can shift very quickly, which may contribute to the larger movements seen in the serve-based probability estimates. This explanation should still be treated with caution, since this study does not test in depth whether this correlation is sufficient to prove causation.

The results were also compared with DeepTennis,\cite{lerner2019} an LSTM-based live prediction model trained on Grand Slam data from 2011--2019. DeepTennis reported an overall accuracy of 79.5\%, with accuracies of 73.2\%, 84.9\%, and 89.2\% at 25\%, 50\%, and 75\% match progress, respectively.

Although DeepTennis uses very similar evaluation methods, its results are not directly comparable to the main results of this study because the data were split differently over time. DeepTennis uses 2017 for development and 2014 for testing, so the evaluation is not fully chronological. This raises a major red flag, as the dataset includes matches from both before and after 2014, allowing the possibility for later-season information to have been available during training. By contrast, this study uses a strictly chronological split, with all training and validation data coming before the test period.

To examine the effect of this difference, the models in this study were also evaluated under the same non-chronological testing setup. Under this alternative split, multiple models matched or exceeded the performance reported by DeepTennis. Table~\ref{tab:accuracy-2011-2019} presents this comparison:

\begin{table}[H]
\centering
\caption{Accuracy under the 2011--2019 non-chronological split; HGBM denotes the histogram gradient-boosting model, and Trace reaches 81.30\% across all points on this comparison split. Source: authors' model evaluation.\cite{xie2026}}
\label{tab:accuracy-2011-2019}
\resizebox{\columnwidth}{!}{%
\begin{tabular}{lcccc}
\toprule
\textbf{Model} & \textbf{25\%} & \textbf{50\%} & \textbf{75\%} & \textbf{All Points} \\
\midrule
Symmetric Markov & 69.24\% & 80.64\% & 87.99\% & 76.73\% \\
Elo-asymmetric Markov & 77.40\% & 83.05\% & 88.70\% & 81.12\% \\
Serve-shrink Markov & 77.53\% & \textbf{83.84\%} & 88.83\% & 81.29\% \\
HGBM & 71.81\% & 81.50\% & 88.48\% & 77.53\% \\
\textbf{Trace} & \textbf{77.66\%} & \textbf{83.84\%} & \textbf{88.96\%} & \textbf{81.30\%} \\
\bottomrule
\end{tabular}
}
\end{table}

These comparison results should be interpreted cautiously because the split allows for potential data leakage. However, because DeepTennis was the strongest comparable model found, the comparison still suggests that Trace is competitive with prior live-tennis prediction approaches.

\section{Limitations}

This project has several limitations. First, the models in this study mainly explore service points and features. Future models could incorporate return features and statistics, which could add additional insights into a player's performance.

Another limitation is the lack of surface-specific performance. Players can perform very differently across hard, clay, and grass courts, with Rafael Nadal's success on clay being a well-known example.\cite{bodo2010} The current Elo system does not directly account for different types of surfaces, so adding surface-specific Elo ratings or performance statistics could provide a more accurate measure of a player's strength.

Despite the fact that the number of features used for this project is already quite large, there are still several factors that are not represented, including fatigue, injury, weather, handedness, and playing style. These factors could undoubtedly have a sizeable impact on live performance, which could give models a better estimate of a player's overall performance. Although some of these effects may be indirectly reflected by the existing data, giving them a more concrete representation may make it easier for models to incorporate these additional features.

Finally, this project only analyzed Grand Slam matches. Since Grand Slams represent only a small fraction of total professional tennis tournaments, the results may not generalize well to lower-level tour events. Hence, adding ATP and WTA tour-level events to the existing dataset could expand the usefulness of these models.

\section{Conclusion}

This study developed a live tennis win-probability model that combines Markov score recursion with pre-match and live match information. Five models were tested, including a symmetric Markov baseline, an Elo-asymmetric Markov model, a serve-shrink model, an HGBM baseline, and Trace. Trace combined features from the other models, making it the main focus of the evaluation. Trace produced the greatest accuracy among the five models in the primary evaluation, reaching 77.84\% across all points. The results of the other four models reveal many interesting patterns between the uses of different sources of information. These results demonstrate that each source of information is valuable for tennis models. In particular, Elo ratings provide useful information on player strength, live serve statistics improve probability estimates, and the HGBM captures additional patterns that are not represented directly by the Markov models.

While much tennis forecasting focuses on pre-match prediction, this study demonstrates the practicality of developing a live-match prediction model.

\section{Future Work}

The main direction of work is to build Trace into a real-time prediction system. As an example of how this would work, the system would receive point-by-point data from a live match, update the score state and live features, and then provide a new match-win probability after every point.

More testing is necessary to ensure that the model remains accurate and efficient when applied to real-time tennis matches.

\section{Implementation Notes}

The statistical analysis was implemented in \href{https://www.python.org/}{Python 3.12.3} using \href{https://numpy.org/}{NumPy 2.5.0}, \href{https://pandas.pydata.org/}{pandas 3.0.3}, \href{https://scikit-learn.org/}{scikit-learn 1.9.0}, \href{https://xgboost.ai/}{XGBoost 3.3.0}, and \href{https://lightgbm.readthedocs.io/}{LightGBM 4.6.0}. Figures 2--4 were generated with \href{https://matplotlib.org/}{Matplotlib 3.11.0}, and Figure 1 was prepared with \href{https://ctan.org/pkg/pgf}{TikZ/PGF in LaTeX}.

Code for the full pipeline is available in the \href{https://github.com/cx-57/live-tennis-research}{project repository (https://github.com/cx-57/live-tennis-research)}.

\section{Acknowledgements}

This project uses public tennis datasets maintained by Jeff Sackmann, including Grand Slam point-by-point data and ATP/WTA match results.

\end{document}